\documentclass[final]{ceurart}

\usepackage{booktabs}
\usepackage{multirow} 
\usepackage{amsmath,amssymb,amsfonts}
\usepackage{amsthm}
\usepackage{mathrsfs}
\usepackage{algorithmic}
\usepackage{graphicx}
\usepackage{xurl}
\usepackage{siunitx}
\def\BibTeX{{\rm B\kern-.05em{\sc i\kern-.025em b}\kern-.08em
 T\kern-.1667em\lower.7ex\hbox{E}\kern-.125emX}}

\usepackage{listings}
\usepackage{amssymb}
\usepackage{pifont}
\newcommand{\cmark}{\ding{51}}
\newcommand{\xmark}{\ding{55}}
\DeclareUnicodeCharacter{FE0F}{}

\begin{document}

\copyrightyear{2024}
\copyrightclause{Copyright ©️2024 for this paper by its authors. Use permitted under Creative Commons License Attribution 4.0 International (CC BY 4.0)}

\conference{3rd AIxIA Workshop on Artificial Intelligence For Healthcare and 5th Data4SmartHealth co-located with the 23rd Internarional Conference of the Italian Association for Artificial Intelligence,
 November 25--28, 2024, Bolzano, Italy}

\title{Detecting anxiety and depression in dialogues: a multi-label and explainable approach}

\author[1]{Francisco de Arriba-Pérez}[%
orcid=0000-0002-1140-679X,
email=farriba@gti.uvigo.es,
url=https://franciscodearribaperez.ddnsfree.com,
]
\fnmark[1]

\author[1]{Silvia García-Méndez}[%
orcid=0000-0003-0533-1303,
email=sgarcia@gti.uvigo.es,
url=https://silviagarciamendez.ddnsfree.com,
]
\cormark[1]
\fnmark[1]

\address[1]{Information Technologies Group, atlanTTic, University of Vigo, Vigo, Spain}

\cortext[1]{Corresponding author.}
\fntext[1]{These authors contributed equally.}

\begin{abstract}
Anxiety and depression are the most common mental health issues worldwide, affecting a non-negligible part of the population. Accordingly, stakeholders, including governments' health systems, are developing new strategies to promote early detection and prevention from a holistic perspective (\textit{i.e.}, addressing several disorders simultaneously). In this work, an entirely novel system for the multi-label classification of anxiety and depression is proposed. The input data consists of dialogues from user interactions with an assistant chatbot. Another relevant contribution lies in using Large Language Models (\textsc{llm}s) for feature extraction, provided the complexity and variability of language. The combination of \textsc{llm}s, given their high capability for language understanding, and Machine Learning (\textsc{ml}) models, provided their contextual knowledge about the classification problem thanks to the labeled data, constitute a promising approach towards mental health assessment. To promote the solution's trustworthiness, reliability, and accountability, explainability descriptions of the model's decision are provided in a graphical dashboard. Experimental results on a real dataset attain \SI{90}{\percent} accuracy, improving those in the prior literature. The ultimate objective is to contribute in an accessible and scalable way before formal treatment occurs in the healthcare systems.
\end{abstract}


\maketitle

\section{Introduction}

More than 55 million people in the United States suffer from mental illness as indicated by the National Institutes of Health (\textsc{nih}, 2023)\footnote{Available at \url{https://www.usa.edu/blog/mental-health-statistics}, October 2024.}. More in detail, the most common mental conditions are anxiety (\SI{19.1}{\percent}) and major depression (\SI{8.3}{\percent}). At the global level, \SI{4}{\percent} of the population is affected by anxiety disorder. At the same time, 280 million people worldwide suffer from depression, as stated by the World Health Organization (\textsc{who}, 2023)\footnote{Available at \url{https://www.who.int/news-room/fact-sheets/detail/depression}, \url{https://www.who.int/news-room/fact-sheets/detail/anxiety-disorders}, October 2024.}. However, only \SI{25}{\percent} of people suffering from anxiety receive treatment. A recent report by Forbes\footnote{Available at \url{https://www.forbes.com/health/mind/depression-statistics}, October 2024.} completes this information and indicates that \SI{50}{\percent} of people affected by depression go undiagnosed in the primary care system. 

In this regard, it should be noted that traditional screening methods (\textit{i.e.}, those that rely on subjective and time-consuming interviews composed of binary questions for patients and their families) face several issues \cite{ahmed2024decoding}. Among them, the unreliability of self-reported diagnoses due to bias introduced by subjectivity, intentional concealment, and even inconvenience of the number of questions must be considered, resulting in the latter low rates of diagnosis and interventions \cite{Lee2022}. Another concern is stigma, which prevents treatment seeking and ignorance of the condition \cite{ahmed2024decoding}. Representative examples of these self-reporting methodologies are the Beck Depression Inventory (\textsc{bdi}), the General Health Questionnaire (\textsc{ghq}), the Hamilton Rating Scale for Depression (\textsc{hrsd}), and the Patient Health Questionnaire (\textsc{phq}). Similar to the \textsc{ghq} and the \textsc{phq}, the Depression, Anxiety, and Stress Scale (\textsc{dass}) combines the questionnaires of each factor \cite{saylam2024multitask}. Consideration should also be given to the popular Diagnostic and Statistical Manual of Mental Disorders, fifth edition (\textsc{dsm-5}) published by the American Psychiatric Association.

Provided the severe consequences of anxiety and depression that even increase the risk of suicide \cite{Marwaha2023}, early detection and timely diagnoses are critical. In this regard, language can be a good predictor of mood disorders \cite{tumaliuan2024development}. More in detail, how users engage in a conversation and express themselves is a strong indicator of their mental health state. Accordingly, the arrival of Large Language Models (\textsc{llm}s, \textit{e.g.}, \textsc{gpt-4}\footnote{Available at \url{https://openai.com/index/gpt-4}, October 2024.}, Pa\textsc{lm}\footnote{Available at \url{https://ai.google/discover/palm2}, October 2024.}, and Alpaca\footnote{Available at \url{https://huggingface.co/models?other=alpaca}, October 2024.}) has contributed significantly to health-related topics thanks to their context-learning capabilities, mostly in generative tasks. Specifically, the literature has reported promising performance of these models in three relevant scenarios: (\textit{i}) language comprehension, (\textit{ii}) text generation, and (\textit{iii}) knowledge inference \cite{Wang2023}. Moreover, the potential of these models to leverage large volumes of online data is of great importance for both diagnosis and treatment \cite{chowdhury2024harnessing}. 

Consequently, several pre-trained language models (\textsc{plm}s) and \textsc{llm}s have been deployed for addressing health issues like mental disorders. It is the case of \textsc{llm}ental \cite{nowacki2024llmental}, Mental\textsc{bert} \cite{Ji2022} and Mental\textsc{ll}a\textsc{ma} \cite{yang2024mentallama}. Besides, Psych\textsc{bert} \cite{Vajre2021} is fine-tuned to detect language patterns in behavioral health, mental health, psychiatry, and psychology texts. However, as indicated by previous works \cite{ahmed2024decoding}, their performance in specific classification problems with task-specific data like anxiety and depression is still immature when used as final solutions (\textit{i.e.}, in zero-shot/few-shot learning or with limited fine-tuning). This is due to the poor detection of nuances and task-specific patterns essential for accurate detection. Similarly, the \textsc{plm}s exhibit limited generalization and low multitask robustness \cite{Wang2022,Moor2023}. Another key limitation is their low interpretability, which prevents their practical use beyond academic research \cite{yang2024mentallama}. 

Summing up, using \textsc{plm}s or directly \textsc{llm}s in zero-shot settings is the prominent approach \cite{liu2024leveraging}. Regarding the experimental data, most researchers use social media \cite{xu2024mental}. In recent years, there has even been an increasing interest in detecting mental health states with tracking devices, which results from the growing importance that modern society places on mental well-being \cite{Gomes2023}. Regardless of the approach, the aspect in which most of the researchers meet is the necessity to provide interpretable results along with explainable descriptions of the rationale of the machine-based solutions, of uttermost importance in the healthcare field provided their direct impact in the decision-making of clinicians and thus, the patient's well-being. In this regard, eXplainable Artificial Intelligence (\textsc{xai}) comprises post hoc and self-explanatory techniques. While the former posthoc alternatives aim to explain the prediction of black-box classification models like the popular explanatory model-agnostic tools (\textit{i.e.} those that combine local linear and random models, like \textsc{lime} and \textsc{shap} \cite{salih2024perspective} to approximate feature importance weights with regression and game theory), the self-explanatory approach relies on intrinsically interpretable models that can provide explanations along with the predictions \cite{liu2024leveraging}. However, feature importance methods like \textsc{lime} and \textsc{shap} only provide the weight of the selected features without considering the interactions among the features and are low intuitive for end users \cite{Coroama2021}. In this regard, a major regulatory milestone in the \textsc{ai} field was materialized with the Artificial Intelligence Act (\textsc{aia}). The final text pays particular attention to interpretability, the right of end users to receive clear explanations, and the disclosure of the use of \textsc{ai} in human interactions \cite{nannini2024operationalizing}.

Given the safety-critical nature of these conditions, our solution must provide high accuracy and explainability to promote trust among the end users and professionals. Accordingly, we combine the traditional Machine Learning (\textsc{ml}) models (which can offer higher accuracy but lack explainability) operating in a multi-label setting with \textsc{llm}s (which are intrinsically explicable but lack specific downstream knowledge). Note that in our approach, \textsc{llm}s are leveraged to extract users’ expert features related to anxiety and depression by detecting linguistic patterns and language usage, taking advantage of their understanding capabilities. Specifically, relying on \textsc{llm}s solely as part of the feature engineering module to extract user-level knowledge; we tackle the hallucination problem, that is, those predictions that, even seem correct, present underlying misconceptions due to the absence of a comprehensive understanding of the problem and expert data. Furthermore, accurate diagnosis requires formal clinical knowledge \cite{tumaliuan2024development}. Consequently, in this work, we understand the necessity of leveraging formal medical knowledge into machine-based solutions (\textit{i.e.}, by incorporating transparent assessment based on official methods, scales, and standards). Hence, we used formal clinical scales for anxiety and depression to label the experimental data. We also acknowledge the limitations of using social media data. Thus, we exploit free dialogues with a conversational assistant. Compared to free dialogues, clinical questionnaires limit the users' ability to express their state freely. Our ultimate objective is to perform an on-demand and scalable assessment of anxiety and depression before formal clinical screening in the healthcare systems. Note that intentional concealment is reduced in our study since the tests are embedded in the dialogues, and the questions are adapted accordingly.

The remainder of this manuscript is divided into the following sections. Section \ref{sec:related_work} summarizes the key prior works on anxiety and depression detection using \textsc{plm}s and \textsc{llm}s, paying particular attention to multi-label approaches and those that provide explainability. Section \ref{sec:methodology} details our system architecture, while Section \ref{sec:results} shows the results obtained with our methodology and compares them with other works in the state of the art. Finally, Section \ref{sec:conclusion} does the main conclusions of this work and proposes future research.

\section{Literature review}
\label{sec:related_work}

Traditional \textsc{ml}, deep learning, and Natural Language Processing (\textsc{nlp}) techniques have been used in the literature for mental health assessment \cite{saylam2024multitask}. The most recent works involve word embedding with transformed-based models (also known as \textsc{plm}s) to take advantage of contextual data \cite{Ren2021,rahman2024depressionemo}. However, scant research is available in state of art related to using \textsc{llm}s like Chat\textsc{gpt} underlying models; most use them as final classification solutions with limited prompt engineering or fine-tuning.

Regarding the detection of anxiety and depression, many researchers apply a multitask learning perspective (\textit{i.e.}, defining a primary and an auxiliary task), \textit{e.g.}, emotion inference. This is due to the availability of experimental labeled data in terms of emotional content \cite{Levenson2019}. These works sustain that stressed users are more likely to express negative emotions (\textit{e.g.}, anger, fear, and sadness) rather than positive ones (\textit{e.g.}, happiness). This is the case in many works. In this regard, \citet{Qureshi2020} defined emotion classification as the second task, which follows depression as the main task, similar to what \citet{Ghosh2022Bio} proposed. Moreover, the solution developed by \citet{Turcan2021} is another representative example. Notably, the authors applied this approach to stress detection. They explored single-task models that operate similarly to \textsc{bert} and multitask learning with a fine-tuned \textsc{bert} model on emotion detection and stress labels. Finally, they exploited \textsc{lime} for interpretability. Although we agree with the strong relation between emotion load and mental health state, we believe that relying mainly on emotion detection to assess anxiety, depression, or stress may lead to false positive results. Thus, we incorporated this knowledge into the engineered features, using anxiety and depression-labeled data as main tasks jointly in a multi-label setting. 

Moreover, \citet{Ghosh2022} proposed a multitasking framework (not based on \textsc{ml} models) for depression detection, sentiment classification, and emotion recognition. Even if slightly related to our research, the promising results obtained prove the appropriateness of addressing anxiety and depression simultaneously, given the strong link between both mental health conditions. Conversely, \citet{Sarkar2022} developed a multitask learning solution with a data-sharing mechanism, providing the relation between anxiety and depression. The authors used word embedding models like \textsc{bert} for feature engineering to feed traditional \textsc{ml} models, similar to our work but without the advantage in terms of explainability that \textsc{llm}s provide. Alike to the work by \citet{Sarkar2022} is the more recent proposal by \citet{park2024probability}. Additionally, \citet{Ilias2023} defined a multitask learning framework in which depression and stress detection are the main and auxiliary tasks, respectively, using social media data. Note that two datasets gathered and labeled in different conditions are used. The first proposed approach encompasses a \textsc{bert}-based layer shared for both tasks, primary and auxiliary, followed by separate \textsc{bert}-based encoder layers. In contrast, the second approach derives from the first but exploits weighting layers by attention fusion networks. However, no hyperparameter tuning was performed due to limited access to computational resources. Explainability was not provided either.

Despite the strong relation between anxiety or stress and depression, few studies address the joint assessment of several conditions \cite{Ilias2023}. In this regard, \citet{Lee2022} focused on geriatric (\textit{i.e.}, experimental data from mild cognitive impairment patients) anxiety and depression detection by exploiting low-cost activity trackers. Regarding the multi-label classification approach followed, the authors applied the binary relevance method. That is, unlike in our work, they used two single-label classifiers for anxiety and depression, respectively, which is a more straightforward way of approaching the problem. However, accuracy may be compromised since the solution does not consider the correlation between labels. As in our work, they include questionnaire-based features from the geriatric anxiety inventory (\textsc{gai}) and the geriatric depression scale (\textsc{gds}). In addition, \citet{Park2023} also integrated the \textsc{dsm-5} diagnostic criteria into their predictive methodology, which is based on a variant of the \textsc{bert} model. Similarly, \citet{Souza2022} proposed a stacking solution with two single-binary classifiers for anxiety, depression, and their comorbidity leveraging social media data. Note that the authors used \textsc{shap} for interpretability. 

Some authors exploited the already mentioned \textsc{plm}s. It is the case of \citet{ahmed2024decoding} who proposed a transformer-based architecture for multi-class depression detection (\textit{i.e.}, in severity levels: absent, mild, moderate, and severe). After text processing, different variants of the \textsc{bert} model are used for classification. The final result is obtained following a voting approach. The authors applied \textsc{lime} to provide interpretability to the solution. Ultimately, the proposed system was compared with Chat\textsc{gpt} (gpt-3.5-turbo model, non-fine-tuned), which attained poor performance. Related to our work, \citet{chowdhury2024harnessing} studied early depression detection from social media data using \textsc{llm}s (\textit{i.e.}, \textsc{gpt-4}), deep learning (\textit{e.g.}, \textsc{lstm}) and transformer models (\textit{e.g.}, \textsc{bert}). However, the authors' approach to explanability is to provide feature-level interpretability. More recently, \citet{Ilias2024} developed a transformer-based solution for stress and depression detection from social media data. Extra-linguistic information is introduced to the \textsc{bert} and Mental\textsc{bert} models. However, the solution does not approach the detection task simultaneously, as in our work, which is much more challenging. Conversely, experiments were performed with datasets for binary classification of stress and depression, respectively, and a multi-class (\textit{i.e.}, with different severity levels) depression dataset.

When it comes to the application of \textsc{llm}s, \citet{Wang2023} leveraged a fine-tuned version of Chat\textsc{gpt} to detect depression. To ensure accurate predictions, the authors proposed a knowledge-enhanced pre-training scheme with emotion analysis capabilities and human feedback. Moreover, \citet{liu2024leveraging} used Chat\textsc{gpt} for data collection along with manually created psychology data that feed \textsc{bert} and Ro\textsc{bert}a models for depression detection. Regarding interpretability, \textsc{shap} was exploited. Similarly, \citet{ohse2024zero} investigated several \textsc{plm}s and \textsc{llm}s (\textit{e.g.}, \textsc{bert}, \textsc{gpt-4}, \textsc{ll}a\textsc{ma}) for depression assessment using clinical interviews as experimental data. The authors exploited the models following the zero-shot paradigm without fine-tuning or prompt engineering. Despite being a relevant study to endorse the applicability to the mental health field of these models, the authors did not exploit their full potential, as already mentioned with the lack of tuning and also regarding explainability. Furthermore, \citet{wang2024explainable} proposed a solution that searches for depression-related texts from the \textsc{bdi} questionnaire. Then, \textsc{llm}s are used to fill the latter survey using user data from social media to infer their mental state. Ultimately, \citet{xu2024mental} evaluated different \textsc{llm}s (\textit{e.g.}, Alapaca, \textsc{ll}a\textsc{ma}, \textsc{gpt-4}) for mental health classification (binary and multi-class prediction for stress, depression and suicide) from social media data exploiting prompt engineering. Note that this work differs from ours in the absence of the multi-label setting and explainable capabilities.

\subsection{Research contributions}

Table \ref{tab:comparison} shows the most closely related solutions to easily compare and assess our contributions. To the best of our knowledge, our work is the first to apply \textsc{llm}s to extract users’ expert features related to anxiety and depression. By this means, we can detect linguistic patterns and language usage, using the comprehension capabilities of \textsc{llm}s without sacrificing explainability. Moreover, another relevant contribution is combining traditional \textsc{ml} models in a multi-label setting, which can offer higher accuracy. Consideration should also be given to integrating formal clinical knowledge through standard tests used for data labeling. More in detail, experimental data consists of a free conversation between patients and a conversational chatbot, despite the popularity of social media data for anxiety and depression detection and the rigidity of self-reporting questionnaires. Ultimately, an explainability dashboard describes the most relevant data that leads to the classification decision and its confidence.

Summing up, the main contributions of the proposed solution for the field are:

\begin{itemize}
 \item A multi-label framework able to predict jointly anxiety and depression.
 \item The use of \textsc{llm}s to extract high-level reasoning features used to train the \textsc{ml} models.
 \item The explainability dashboard which promotes trust and makes the solution accountable and reliable.
\end{itemize}

\begin{table*}[!htbp]
\centering
\caption{Comparison of related solutions.}
\label{tab:comparison}
\begin{tabular}{lccccc}
\toprule
\textbf{Authorship} & \textbf{Approach} & \bf Data & \textbf{Explainability}\\
\midrule

\citet{Lee2022} & Multi-label \textsc{ml} & Tracker and clinical data & \xmark\\

\citet{Wang2023} & \textsc{llm}s & General and depression related text & \xmark\\

\citet{ohse2024zero} & \textsc{plm}s/\textsc{llm}s & Clinical interviews & \xmark\\

\citet{xu2024mental} & \textsc{llm}s & Social media & \xmark\\

\midrule

Our proposal & Multi-label \textsc{ml} + \textsc{llm}s & Dialogues + clinical data & \cmark\\
 
\bottomrule
\end{tabular}
\end{table*}

\section{Methodology}
\label{sec:methodology}

\begin{figure*}[!htpb]
 \centering
 \includegraphics[scale=0.12]{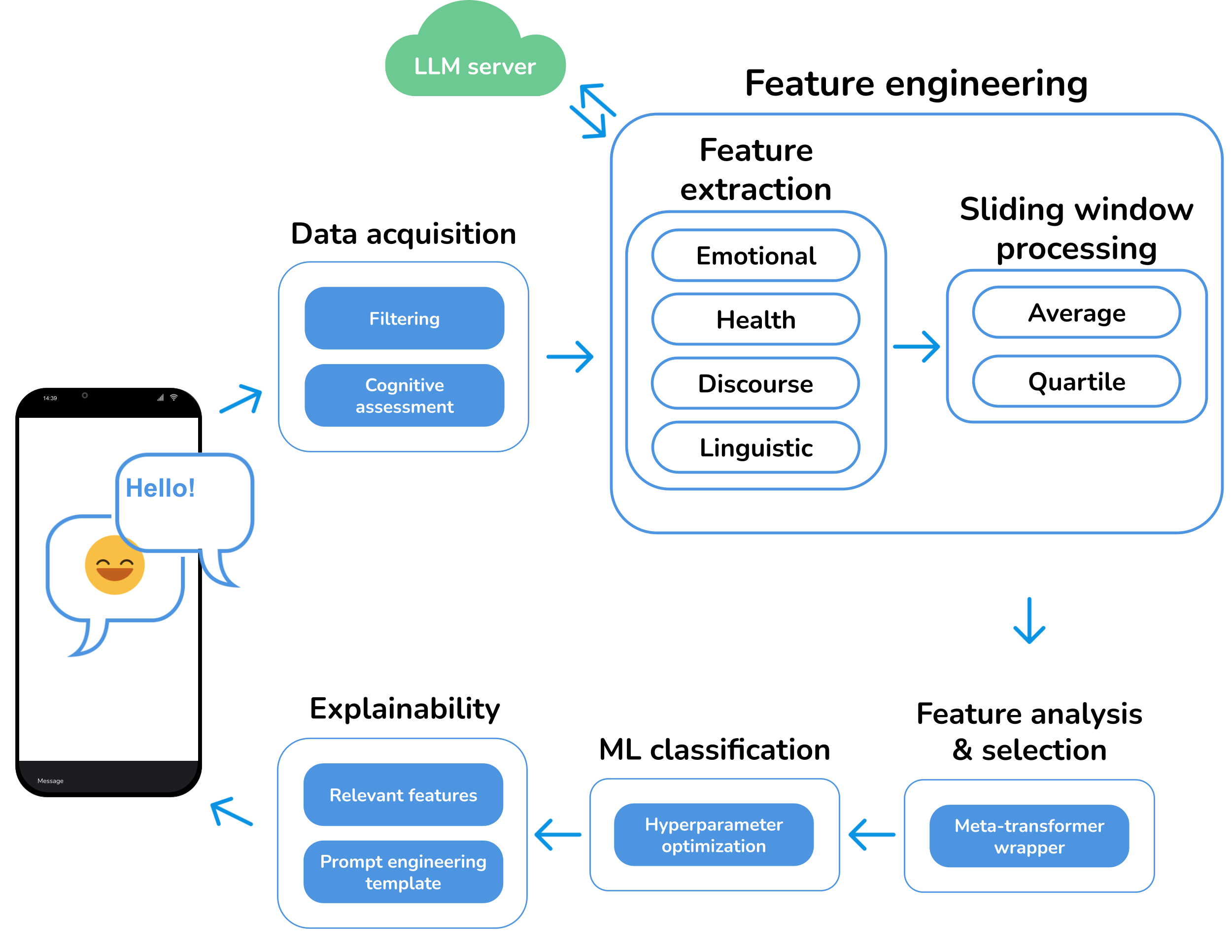}
 \caption{System scheme.}
 \label{fig:scheme}
\end{figure*}

Figure \ref{fig:scheme} shows the scheme of our proposal. It is composed of (\textit{i}) data acquisition to gather and filter data; (\textit{ii}) feature engineering to generate new features using prompt engineering through an \textsc{llm} model and applying sliding window treatment; (\textit{iii}) feature analysis and selection to remove no-relevant features. Moreover, the (\textit{iv}) classification module evaluates our multi-label solution. Finally, (\textit{v}) the explainability module generates a natural language explanation of the classifier predictions leveraging a prompt engineering template.

\subsection{Data acquisition}
\label{sec:data_acquisition}

The experimental dataset is composed of conversations with \textit{Celia} chatbot\footnote{Available at \url{https://celiatecuida.com/en/home_en}, October 2024.}. This chatbot establishes an entertaining and engaging dialogue with end users, including fun facts about the conversation topics. Moreover, every 3 months, the chatbot uses the standard questionnaires presented in the Spanish versions of the Goldberg Anxiety and Depression Scales (\textsc{gads}) and the Yesavage Geriatric Depression Scale (\textsc{ygds}) to assess the cognitive state of the user. These questions are embedded during the conversation flow. The latter data is used as the label of the user (\textit{i.e.}, absence or presence of anxiety and depression) for the supervised learning stage.

\subsection{Feature engineering}
\label{sec:feature_engineering}

The solution combines feature generation based on prompt engineering with a sliding window strategy to consider the history of past sessions\footnote{A session is a complete dialogue with the end user until they decide to stop the conversation.}. Table \ref{tab:features} shows the features engineered that can verse on the cognitive state of the end user (\textit{i.e.}, their emotional well-being or health condition) or on the dialogue itself (\textit{i.e.}, the discoursive and linguistic characteristics of the conversation with the chatbot). These features are calculated using an \textsc{llm} and prompt engineering, and their values range from 0.0 to 1.0\footnote{Polarity takes integer values: 0 for negative, 1 for neutral, and 2 for positive.}.

Each generated feature is expanded with four new statistical features (average, and the three quartiles $Q_1$, $Q_2$, and $Q_3$). For this purpose, a sliding window is performed with the last 30 sessions (see Equation \eqref{eq:feature_engineering}, where $n$ is the number of sessions and $X[n]$ is the historical feature in the last $n$ sessions. Consequently, $Y[n]$ is the ordered version of $X[n]$.

\begin{equation}\label{eq:feature_engineering}
\begin{split}
\forall n \in \{1 ... \infty\}\\ \\
X[n] = \{ x[0],\ldots,x[n]\}. \\
Y[n] = \{y_0[n], y_1[n],\ldots,y_{n-1}[n]\} \mid  y_0[n]\leq y_1[n]\leq\ldots\leq y_{n-1}[n], \\
\mbox{where} ; \forall x \in X[n], \; x \in Y[n]. \\ \\
avg[n]=\frac{1}{n}\sum_{i=0}^{n} y_i [n]\\
Q_{1}[n]=y_{\lfloor\frac{1}{4}n\rceil}[n] \\
Q_{2}[n]=y_{\lfloor\frac{2}{4}n\rceil} [n]\\
Q_{3}[n]=y_{\lfloor\frac{3}{4}n\rceil} [n]\\
\end{split}
\end{equation}

\begin{table*}[!htbp]
\centering
\caption{\label{tab:features}Features engineered for anxiety an depression detection.}
\begin{tabular}{llllp{6cm}}
\toprule
\bf Group & \bf Category & \bf ID & \bf Name & \bf Description\\\hline

\multirow{7}{*}{Cognitive state} 
&\multirow{5}{*}{Emotional} 
& 1 & Insecurity & Level of insecurity or fear.\\
&& 2 & Loneliness & Level of isolation.\\
&& 3 & Negative emotion & Level of negative charge.\\
&& 4 & Positive emotion & Level of positive charge.\\
&& 5 & Sadness & Level of sorrow.\\

\cmidrule(lr){2-5}
 
&\multirow{2}{*}{Health} 
& 6 & Anguish & Level of distress.\\
&& 7 & Health issues & Level of health concerns.\\

\midrule

\multirow{9}{*}{Content} 
&\multirow{5}{*}{Discourse} 
& 8 & Catastrophic terms & Level of use of disaster charge.\\
&& 9 & Emphasized terms & Level of use of emphasis (positive or negative).\\
&& 10 & Repeated concepts & Extent to which the user repeats previous content.\\

\cmidrule(lr){2-5}
&\multirow{4}{*}{Linguistic} 
& 11 & Interjections & Level of use of interjections.\\
&& 12 & Negative adverbs & Level of use of negative adverbs.\\
&& 13 & Negatives terms & Level of use of negative terms.\\
&& 14 & Polarity & Negative, neutral, positive.\\

\bottomrule
\end{tabular}
\end{table*}

\subsection{Feature analysis \& selection}
\label{sec:feature_analysis_selection}

In the cold-start step, the system uses \SI{10}{\percent} of the samples to select the most relevant features. In this first phase, a selector based on a meta-transformer wrapper\footnote{Available at \url{https://scikit-learn.org/1.5/modules/feature_selection.html}, October 2024.} is applied (\textit{i.e.}, following a model-agnostic strategy). In order to select the most significant features, this transformer uses a tree \textsc{ml} classifier that calculates the Mean Decrease in Impurity (\textsc{mdi})\footnote{Available at \url{https://scikit-learn.org/1.5/auto_examples/inspection/plot_permutation_importance.html}, October 2024.} of each feature. Ultimately, the features with a \textsc{mdi} lower than average are discarded.

\subsection{Classification}
\label{sec:classification}

In this study, we solve a multi-label binary classification problem. More in detail, this scenario differs from the binary class in the number of labels assigned. In the latter binary-class scenario, the classifier provides a single label between the two classes in the experimental data (\textit{i.e.}, number of classes is 2, the number of resulting labels is 1). Since ours is a binary problem, we applied the Multi-class Transformation Strategy (\textsc{mts}) \cite{Rivolli2020}. Particularly, in our approach, we group the binary classes into four categories: (none\_none), (none\_depression), (anxiety\_none), and (anxiety\_depression). Note that the multi-label classification is a complex problem because the results may be partially correct, preventing the use of standard \textsc{ml} evaluation metrics. Instead, the metrics described below (micro and macro approach) are computed:

\begin{itemize}

\item \textbf{Exact match ratio} represents the proportion of predictions where both labels are correct (see equation \eqref{eq:em}, where $A$ is the actual label, $P$ is the predicted label, and $n$ is the number of samples).

\begin{equation}\label{eq:em}
 \frac{1}{n} \sum_{i=1}^{n}I(A_i=P_i)
\end{equation}

\item \textbf{Accuracy} is the percentage of correctly predicted labels over the total predicted and actual categories (see equation \eqref{eq:acc}).

\begin{equation}\label{eq:acc}
 \frac{1}{n} \sum_{i=1}^{n}\frac{|A_i \cap P_i|}{|A_i \cup P_i|}
\end{equation}

\item \textbf{Precision} is the percentage of correctly predicted labels over predicted labels (see equation \eqref{eq:precision}).

\begin{equation}\label{eq:precision}
 \frac{1}{n} \sum_{i=1}^{n}\frac{|A_i \cap P_i|}{|P_i|}
\end{equation}

\item \textbf{Recall} is the percentage of correctly predicted labels over actual labels (see equation \eqref{eq:recall}). 

\begin{equation}\label{eq:recall}
 \frac{1}{n} \sum_{i=1}^{n}\frac{|A_i \cap P_i|}{|A_i|}
\end{equation}

\item \textbf{Hamming Loss} (\textsc{hl}) calculates the incorrectly predicted labels. In our binary problem, it complements the accuracy (see equation \eqref{eq:hl}).

\begin{equation}\label{eq:hl}
 1-Accuracy
\end{equation}

\end{itemize}

Our solution exploits the Naive Bayes (\textsc{nb}), Decision Tree (\textsc{dt}), and Random Forest (\textsc{rf}) classifiers widely used in the literature to solve similar classification problems \cite{Lee2022,Ghosh2022Bio,Sarkar2022}. We analyze two different scenarios. Scenario 1 uses all user sessions to summarize the analysis, while scenario 2 evaluates the behavior of the system by reducing the number of samples and selecting 2 out of 3 entries.

\subsection{Explainability}
\label{sec:explainability}

Our system explains the predictions obtained by leveraging an \textsc{llm} with a prompt engineering template. This approach creates an explanation of the predicted majority category every 7 sessions. For this purpose and to limit the computational load of the explainability module, the most representative statistics of the features in Table \ref{tab:features} (\textit{i.e.}, the average and the $Q_2$) are considered. Moreover, the conversations of the last two sessions are also sent to the model with the predicted category for interpretability purposes. In addition to promoting trust among end users and clinicians and the accountability and reliability of the solution, this information can be exploited to recommend formal assessment in the primary care health system or treatment to prevent anxiety and depression.

\section{Results and discussion}
\label{sec:results}

This section explains the experimental dataset used and the results obtained. The analysis was conducted on a computer with the following specifications:

\begin{itemize}
 \item \textbf{Operating System (\textsc{os})}: Ubuntu 18.04.2 LTS 64 bits
 \item \textbf{Processor}: Intel\@Core i9-10900K \SI{2.80}{\giga\hertz}
 \item \textbf{RAM}: \SI{96}{\giga\byte} DDR4 
 \item \textbf{Disk storage}: \SI{480}{\giga\byte} NVME + \SI{500}{\giga\byte} SSD
\end{itemize}

\subsection{Experimental data}

The dataset\footnote{The experimental dataset is available on request from the authors.} contains the complete conversations between voluntary users and the \textit{Celia} chatbot from 16 May 2023 to 9 October 2024. Notably, it comprises \num{2186} user sessions, 32 users, and an average of 68 sessions per user. Moreover, each session comprises an average of 26 interactions and 157 words per session. Table \ref{tab:dataset_distribution} shows the distribution of sessions by categories. More in detail, most cases are concentrated in people without any pathology, and the presence of depression overlaps with anxiety, reducing the number of isolated cases of the former. This increases the difficulty of the classification problem, given the imbalance of experimental data, even more so in the multi-label scenario.

\begin{table}[!htpb]\centering
\caption{Experimental dataset of anxiety and depression.}
\label{tab:dataset_distribution}
\begin{tabular}{llc}\toprule
\textbf{Anxiety} & \textbf{Depression} & \textbf{\textbf{Number of sessions}}\\
\midrule
Absent & Absent & 1287 \\
Present & Absent & 542 \\
Absent & Present & 29\\
Present & Present & 328\\
\midrule
\multicolumn{2}{c}{Total} & 2186\\
\bottomrule
\end{tabular}
\end{table}

\subsection{Data acquisition}
\label{sec:data_acquisition_result}

To motivate a new conversation with \textit{Celia}, the assistant sends notifications by email and shows reminders to the users. In this line, to detect the end of a session, the user must say goodbye, or this is automatically finished after 3 minutes of inactivity. Those sessions with 5 or fewer human interventions are discarded to ensure that a significant amount of data enters the anxiety and depression detection system. In addition, as described in Section \ref{sec:data_acquisition}, every 3 months, the samples are re-labeled on anxiety and depression using the standard questionnaires \textsc{gads} and \textsc{ygds}.

\subsection{Feature engineering}
\label{sec:feature_engineering_results}

The features described in Table \ref{tab:features} are generated using the \texttt{\textsc{gpt}-4o-mini}\footnote{Available at \url{https://platform.openai.com/docs/models/gpt-4o-mini}, October 2024.} model. It can be accessed by sending requests to Open\textsc{ai} \textsc{api}\footnote{Available at \url{https://openai.com/api}, October 2024.}, using the prompt in Listing \ref{lst:prompt_gpt}. Each request contains the text of the complete session and the temperature parameter set to 0. This removes the randomness and ensures that the model provides the same response to the same input content. 

For each of the 14 features, the average and the three quartiles $Q_1$, $Q_2$, and $ Q_3$ are calculated. In total, 56 features are used in this multi-label problem. Once calculated, the values are rounded to 2 digits, and those with the same values are discarded.

\begin{lstlisting}[frame=single,caption={Feature engineering using \textsc{gpt}-4o-mini.}, label={lst:prompt_gpt},emphstyle=\textbf,escapechar=ä]
This is a conversation between a bot and a person. It analyzes the interactions of 
the person as a whole. Please indicate on a range from 0.0 to 1.0 the level of
insecurity, loneliness, negative emotion, positive emotion, sadness, anguish, 
health issues, and the use of catastrophic terms, exaggerated terms, repeated
concepts, interjections, negative adverbs, negative terms, and polarity. 
Following this JSON format. Do not add any textual explanation:
{insecurity:0,
loneliness:0,
negative_emotion:0,
positive_emotion:0,
sadness:0,
anguish:0,
health_issues:0,
catastrophic_terms:0,
exaggerated_terms:0,
repeated_concepts:0,
interjections:0,
negative_adverbs:0,
negatives_terms:0,
polarity:0 (0 negative, 1 neutral, 2 positive)}
<session>
\end{lstlisting}

\subsection{Feature analysis \& selection}
\label{sec:feature_analysis_selection_results}

Our approach uses the \texttt{SelectFromModel}\footnote{Available at \url{https://scikit-learn.org/stable/modules/generated/sklearn.feature_selection.SelectFromModel.html}, October 2024.} library of \texttt{scikit-learn} in combination with the \textsc{rf} classifier\footnote{Available at \url{https://scikit-learn.org/stable/modules/generated/sklearn.ensemble.RandomForestClassifier.html}, October 2024.} to analyze and select the most relevant features. This analysis is performed using the \SI{10}{\percent} of the dataset. Specifically, \SI{39}{\percent} of the original features are selected in scenario 1 and 2.

\subsection{Classification}
\label{sec:classification_results}

We evaluate our approach using the \texttt{scikit-learn} Python library. The \textsc{nb}\footnote{Available at \url{https://scikit-learn.org/stable/modules/generated/sklearn.naive_bayes.GaussianNB.html}, October 2024.}, \textsc{dt}\footnote{Available at \url{https://scikit-learn.org/stable/modules/generated/sklearn.tree.DecisionTreeClassifier.html}, October 2024.} and \textsc{rf} models are selected.

The hyperparameters of these classifiers are optimized using the \texttt{GridSearchCV}\footnote{Available at \url{https://scikit-learn.org/stable/modules/generated/sklearn.model_selection.GridSearchCV.html}, October 2024.} method. We use a $10$-fold cross-validation to evaluate the configuration that offers the best accuracy value over the \SI{10}{\percent} of the experimental dataset. Listings \ref{nb_conf}, \ref{dt_conf} and \ref{rf_conf} contain the parameter sets for \textsc{nb}, \textsc{dt}, and \textsc{rf}, respectively. The values selected for scenarios 1 and 2 are shown below.

\textbf{Scenario 1:}
\begin{itemize}
\item \textsc{nb}: {\tt var\_smoothing} = 1e-05
\item \textsc{dt}: {\tt splitter} = best,
 {\tt max\_features} = None,
 {\tt max\_depth} = 100,
 {\tt min\_samples\_split} = 0.001,
 {\tt min\_samples\_leaf} = 0.001,
 {\tt criterion} = entropy
\item \textsc{rf}: {\tt n\_estimators} = 200,
 {\tt max\_features} = sqrt,
 {\tt max\_depth} = 10,
 {\tt min\_samples\_split} = 2,
 {\tt min\_samples\_leaf} = 1,
 {\tt criterion} = gini
\end{itemize}

\textbf{Scenario 2:}
\begin{itemize}

\item \textsc{nb}: {\tt var\_smoothing} = 1e-05

\item \textsc{dt}: {\tt splitter} = random,
 {\tt max\_features} = sqrt,
 {\tt max\_depth} = 100,
 {\tt min\_samples\_split} = 0.001,
 {\tt min\_samples\_leaf} = 0.001,
 {\tt criterion} = entropy

\item \textsc{rf}: {\tt n\_estimators} = 100,
 {\tt max\_features} = None,
 {\tt max\_depth} = 5,
 {\tt min\_samples\_split} = 2,
 {\tt min\_samples\_leaf} = 1,
 {\tt criterion} = entropy
 
\end{itemize}

\begin{lstlisting}[frame=single,caption={\textsc{nb} hyperparameter configuration.},label={nb_conf},emphstyle=\textbf,escapechar=ä]
var_smoothing : [1e-9, 1e-5, 1e-1]
\end{lstlisting}

\begin{lstlisting}[frame=single,caption={\textsc{dt} hyperparameter configuration.},label={dt_conf},emphstyle=\textbf,escapechar=ä]
 splitter : [best, random],
 max_features : [None, sqrt, log2],
 max_depth : [1, 100, None],
 min_samples_split : [0.001, 0.1, 1],
 min_samples_leaf : [0.001, 0.1, 1],
 criterion : [gini, entropy]
\end{lstlisting}

\begin{lstlisting}[frame=single,caption={\textsc{rf} hyperparameter configuration.},label={rf_conf},emphstyle=\textbf,escapechar=ä]
 n_estimators : [100,150,200],
 max_features : [sqrt, log2, None],
 max_depth : [5, 10, 100, None],
 min_samples_split : [2, 5, 10],
 min_samples_leaf : [1, 2, 5],
 criterion : [gini, entropy]
\end{lstlisting}

Table \ref{tab:classifiers_results} shows the classification results obtained. Note that the three alternative models attain promising results, \textsc{rf} the one that attained the best values, all above \SI{80}{\percent} regardless of the scenario. Moreover, the accuracy value is close to \SI{90}{\percent} (\textit{i.e.}, \SI{10}{\percent} for \textsc{hl}) and the exact match value is close to \SI{85}{\percent}. In scenario 1, \textsc{rf} obtains an increase of +\SI{22}{\percent} compared to the exact match of \textsc{nb} and +\SI{10}{\percent} compared to the macro precision of \textsc{dt}. Moreover, in scenario 2, the reduction of the dataset has a slight non-critical effect on the classifiers.

Compared to the most closely related work from the literature, even though they exploit data from low-cost activity trackers, the multi-label approach by \citet{Lee2022} attains similar results. However, as explained in the related work discussion, the authors focused on mild cognitive impairment. Moreover, the multi-label strategy is also different as they applied the binary relevance method with single-label classifiers for anxiety and depression separately, not considering the correlation between these two conditions. In contrast, we leverage the \textsc{mts}. Moreover, the experimental data of this work is limited to 20 samples, while our approach results in a more realistic and transparent assessment using free dialogues.

\begin{table*}[!htbp]
\small
\centering
\caption{\label{tab:classifiers_results}\textsc{ml} classification results.}
\begin{tabular}{ccp{1cm}cccccccc}
 \cmidrule(lr){1-11}
 \multirow{2}{*}{{\bf Scenario}} & \multirow{2}{*}{{\bf Model}} & \multirow{2}{*}{{\bf Acc.}} & \multirow{2}{*}{{\bf Exact match}} & \multicolumn{2}{c}{\bf Precision} & \multicolumn{2}{c}{\bf Recall} & \multicolumn{2}{c}{\bf \textit{F}-measure} & \multirow{2}{*}{\bf \textsc{hl}} \\
 \cmidrule(lr){5-6}
 \cmidrule(lr){7-8}
 \cmidrule(lr){9-10}
 & & & & Macro & Micro & Macro & Micro & Macro & Micro & \\
 \midrule
\multirow{3}{*}{\textbf{1}} 
& \textsc{nb} & 72.71 & 62.95 & 73.40 & 77.77 & 72.49 & 77.42 & 72.91 & 75.97 & 27.29\\
& \textsc{dt} & 83.46 & 76.03 & 83.48 & 87.03 & \bf 84.53 & 87.33 & 83.98 & 85.94 & 16.54\\
& \textsc{rf} & \bf 89.43 & \bf 84.72 & \bf 93.50 & \bf 93.39 & 84.18 & \bf 90.19 & 88.03 & \bf 91.00 & \bf 10.57\\

\midrule
 
\multirow{3}{*}{\textbf{2}} 
& \textsc{nb} & 72.61 & 59.03 & 70.23 & 76.97 & 75.82 & 81.85 & 72.91 & 77.14 & 27.39\\
& \textsc{dt} & 86.62 & 78.52 & 84.69 & 89.84 & \bf 86.96 & \bf 91.49 & 85.81 & 89.32 & 13.38\\
& \textsc{rf} & \bf 88.33 & \bf 84.21 & \bf 92.56 & \bf 91.70 & 83.76 & 89.09 & \bf 87.35 & \bf 89.70 & \bf 11.67\\
 
\bottomrule
\end{tabular}
\end{table*}

\subsection{Explainability}
\label{sec:explainability_results}

\begin{figure}[!htpb]
 \centering
 \includegraphics[scale=0.15]{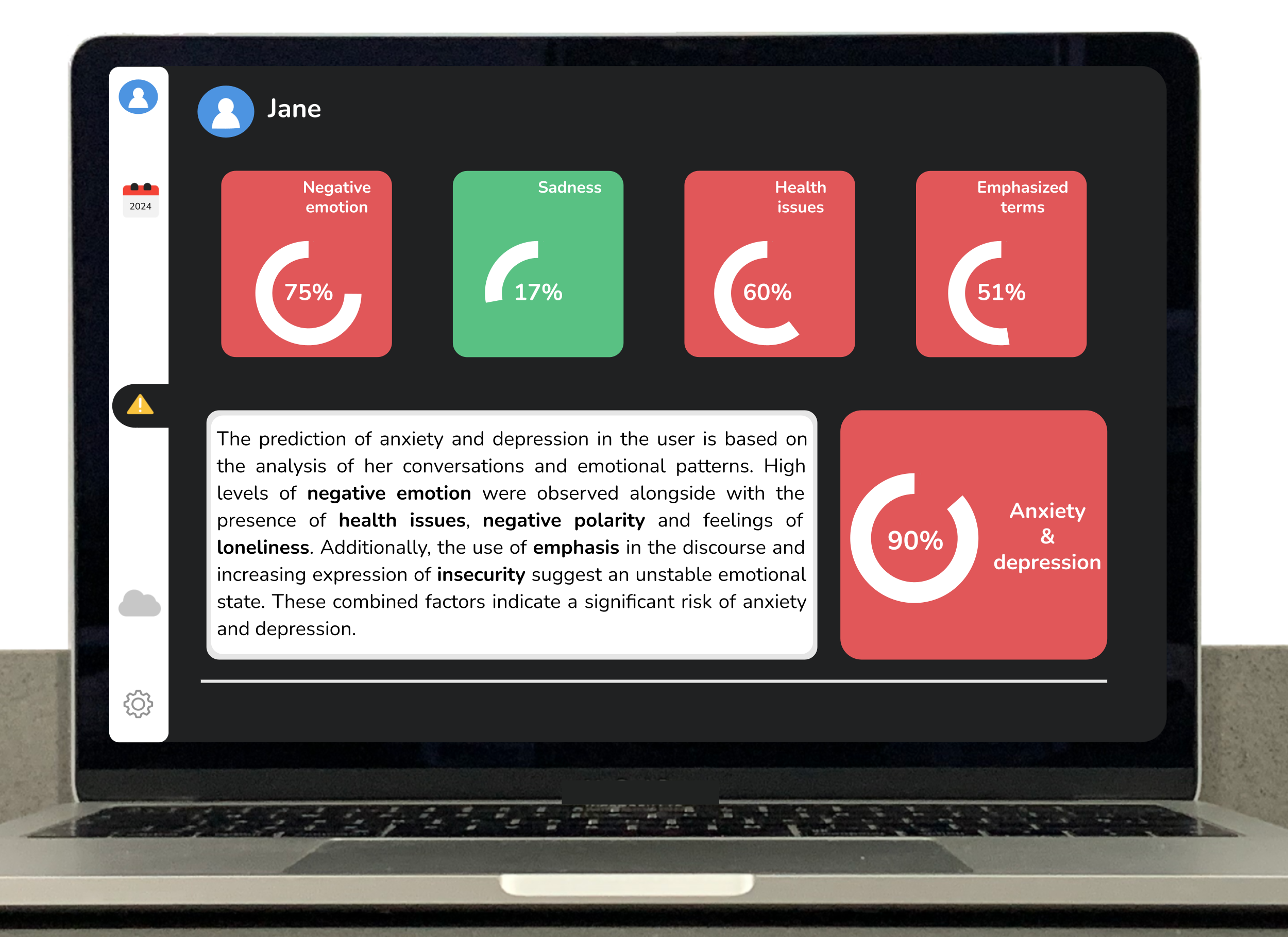}
 \caption{Explainable dashboard.}
 \label{fig:dashboard}
\end{figure}

Figure \ref{fig:dashboard} shows the dashboard accessible to the caregivers, physicians, and end users with the four most relevant features from Table \ref{tab:features} on the top. The boxes are green if the feature values are below \SI{50}{\percent} and red otherwise. At the bottom of the figure, the explanation generated by the \texttt{\textsc{gpt}-4o-mini} model is shown. Our approach uses the prompt engineering template described in Listing \ref{lst:prompt_gpt_expl} filled with the values of the average and the $Q_2$ of the features in Table \ref{tab:features}, the conversations of the last 2 sessions, and the predicted majority category. On the right, the dashboard indicates the prediction and the confidence percentage using \texttt{\small Predict\_Proba} function\footnote{Available at \url{https://scikit-learn.org/stable/modules/generated/sklearn.ensemble.RandomForestClassifier.html}, October 2024.}.

\begin{lstlisting}[frame=single,caption={Prompt for explainability.}, label={lst:prompt_gpt_expl},emphstyle=\textbf,escapechar=ä]

This system analyzes the user's anxiety and depression state using conversations
with a chatbot. The last 30 sessions of the user return the following feature
values:

Average:
    insecurity:X,
    loneliness:X,
    negative_emotion:X,
    positive_emotion:X,
    sadness:X,
    anguish:X,
    health_issues:X,
    catastrophic_terms:X,
    emphasized_terms:X,
    repeated_concepts:X,
    interjections:X,
    negative_adverbs:X,
    negatives_terms:X,
    polarity:X

Q2:
    insecurity:X,
    loneliness:X,
    negative_emotion:X,
    positive_emotion:X,
    sadness:X,
    anguish:X,
    health_issues:X,
    catastrophic_terms:X,
    emphasized_terms:X,
    repeated_concepts:X,
    interjections:X,
    negative_adverbs:X,
    negatives_terms:X,
    polarity:X

Moreover, the last 2 conversations are:

-----------------------------------------
[conversations]
-----------------------------------------

The prediction of our machine learning model is that the user <does not suffer | 
suffers> [anxiety/depression/anxiety and depression].

Generate an exposition of no more than 400 characters in natural language that 
summarizes the reasons why this prediction has been generated by the model based on
the information provided.

\end{lstlisting}

\section{Conclusion}
\label{sec:conclusion}

Given the appalling consequences of anxiety and depression, timely detection of these conditions is of uttermost importance. Traditional screening methods are time-consuming and rely on rigid subjective assessment with interviews and questionnaires. Moreover, despite the strong relationship between anxiety or stress and depression, few studies address the joint assessment of several conditions. 

\textsc{ai}-based solutions have proposed several advantages regarding flexibility, scalability, and personalization. However, their performance in specific classification problems with task-specific data like anxiety and depression is still immature when used as final solutions, as is the case with \textsc{llm}s. Moreover, some solutions lack generalization and multitask robustness, apart from low interpretability, which prevents their practical use beyond academic research. Interpretability and explainability are especially relevant in this field, provided their direct impact on clinicians' decision-making and, thus, the patient's well-being.

In this work, an entirely novel system for the multi-label classification of anxiety and depression is proposed. Another relevant contribution lies in using \textsc{llm}s for feature extraction, which are intrinsically explicable but lack specific downstream knowledge, with \textsc{ml} models operating in a multi-label setting, which can offer higher accuracy but lack explainability. Specifically, relying on \textsc{llm}s solely as part of the feature engineering module to extract user-level knowledge from free dialogues with a conversational assistant, we tackle the hallucination problem. In addition, we leverage formal medical knowledge using clinical scales for anxiety and depression to label the experimental data. Moreover, explainability descriptions of the model's decision are provided in a graphical dashboard along with the confidence of the results to promote the solution's trustworthiness, reliability, and accountability. Experimental results on a real dataset attain \SI{90}{\percent} accuracy, improving those in the prior literature. The ultimate objective is to contribute in an accessible and scalable way before formal treatment occurs in the healthcare systems.

In future work, we plan to evolve the solution to study severity levels of mental health conditions, as well as deploy the system in a real-world setting (\textit{i.e.}, stream-based \textsc{ml}). Another line of work will focus on the analysis of non-verbal and paraverbal data (\textit{e.g.}, voice modulation).

\section*{Declaration of competing interest}

The authors have no competing interests to declare relevant to this article's content.

\section*{Declaration of studies in humans}

This study was carried out following the World Medical Association Declaration of Helsinki.

\begin{acknowledgments}
This work was partially supported by Xunta de Galicia grants ED481B-2022-093 and ED481D 2024/014, Spain.
\end{acknowledgments}

\bibliography{2_bibbliography}

\end{document}